\documentclass[runningheads]{llncs}
\usepackage{amsmath,amssymb,amsfonts}
\usepackage{cite}
\usepackage{algorithmic}
\usepackage{textcomp}
\usepackage{xcolor}
\usepackage{comment}
\usepackage{url}
\usepackage{algorithm}
\usepackage{tikz}
\usetikzlibrary{matrix}
\usepackage{siunitx}
\usepackage{comment}
\usepackage{float}
\usepackage{graphicx, subfigure}
\usepackage{etoolbox}
\usepackage{epstopdf}
\usepackage{setspace}
\usepackage{fancyhdr}
\usepackage{lipsum}
\usepackage{graphicx}
\usepackage[T1]{fontenc}
\usepackage{graphicx}
\begin{document}
\title{Cardiovascular Disease Risk Prediction via Social Media}
%
%
\author{Al Zadid S.B. Habib\inst{1} \and
M.A. Bin Syed\inst{2} \and
M.T. Islam\inst{3} \and
Donald A. Adjeroh\inst{1}}
\authorrunning{AZSB Habib et al.}
%
\institute{Lane Department of Computer Science and Electrical Engineering\and
Department of Industrial and Management Systems Engineering \and
Wadsworth Department of Civil and Environmental Engineering\\
West Virginia University, Morgantown WV 26506, USA\\
\email{\{ah00069, ms00110, mi00018\}@mix.wvu.edu}, donald.adjeroh@mail.wvu.edu}
\maketitle     
\begin{abstract}
Researchers use Twitter and sentiment analysis to predict Cardiovascular Disease (CVD) risk. We developed a new dictionary of CVD-related keywords by analyzing emotions expressed in tweets. Tweets from eighteen US states, including the Appalachian region, were collected. Using the VADER model for sentiment analysis, users were classified as potentially at CVD risk. Machine Learning (ML) models were employed to classify individuals' CVD risk and applied to a CDC dataset with demographic information to make the comparison. Performance evaluation metrics such as Test Accuracy, Precision, Recall, F1 score, Mathew's Correlation Coefficient (MCC), and Cohen's Kappa (CK) score were considered. Results demonstrated that analyzing tweets' emotions surpassed the predictive power of demographic data alone, enabling the identification of individuals at potential risk of developing CVD. This research highlights the potential of Natural Language Processing (NLP) and ML techniques in using tweets to identify individuals with CVD risks, providing an alternative approach to traditional demographic information for public health monitoring.
\keywords{Social Media Analytics  \and Cardiovascular Disease \and Machine Learning \and Deep Learning \and CDC \and CNN-LSTM.}
\end{abstract}
\section{Introduction}
CVD is one of the most significant health concerns related to morbidity and mortality \cite{b1}. For example, heart disease and stroke are the first and third most common fatalities in the US, respectively \cite{b2}. A previous study found that one in three Americans have one or more CVDs \cite{b3}. Psychological traits have been demonstrated to raise the risk for CVD via physiological consequences and unhealthy behaviors \cite{b4}. Social media data now contains a wealth of information about communities’ psychological conditions and behaviors. It is a good platform for people to share personal experiences, seek information, and exchange mental support or sympathy on health issues. Thus, through Twitter, researchers can systematically observe public discourse on health issues, such as CVD \cite{b5}. In essence, it is achieved through NLP by determining the polarity of word data, which is divided into negative, positive, and neutral categories \cite{b6,b7,b8}. Moreover, existing literature supports the hypothesis that there are no significant differences in the proportion of social media users irrespective of psychological characteristics, which nullifies the concern regarding biased estimates.

Convolutional Neural Network (CNN) is a Deep Learning (DL) architecture that extracts features from the input before sending it through filters. Long Short-Term Memory (LSTM) is a type of neural network that attempts to give the Recurrent Neural Network (RNN) a short-term memory that can endure thousands of timesteps. Thus, the Hybrid CNN-LSTM architecture can classify people into high to low-risk categories. Since NLP-based research is still in its infancy, the risk prediction of CVDs with hybrid CNN-LSTM is yet to be explored. Moreover, several ML algorithms like Bernoulli Naive Bayes (BNB), Support Vector Machine (SVM), Logistic Regression (LR), and CatBoost are known to be good with categorical data.

\begin{figure}
\centering     
\subfigure[Selected States]{\label{fig1}\includegraphics[width=40mm]{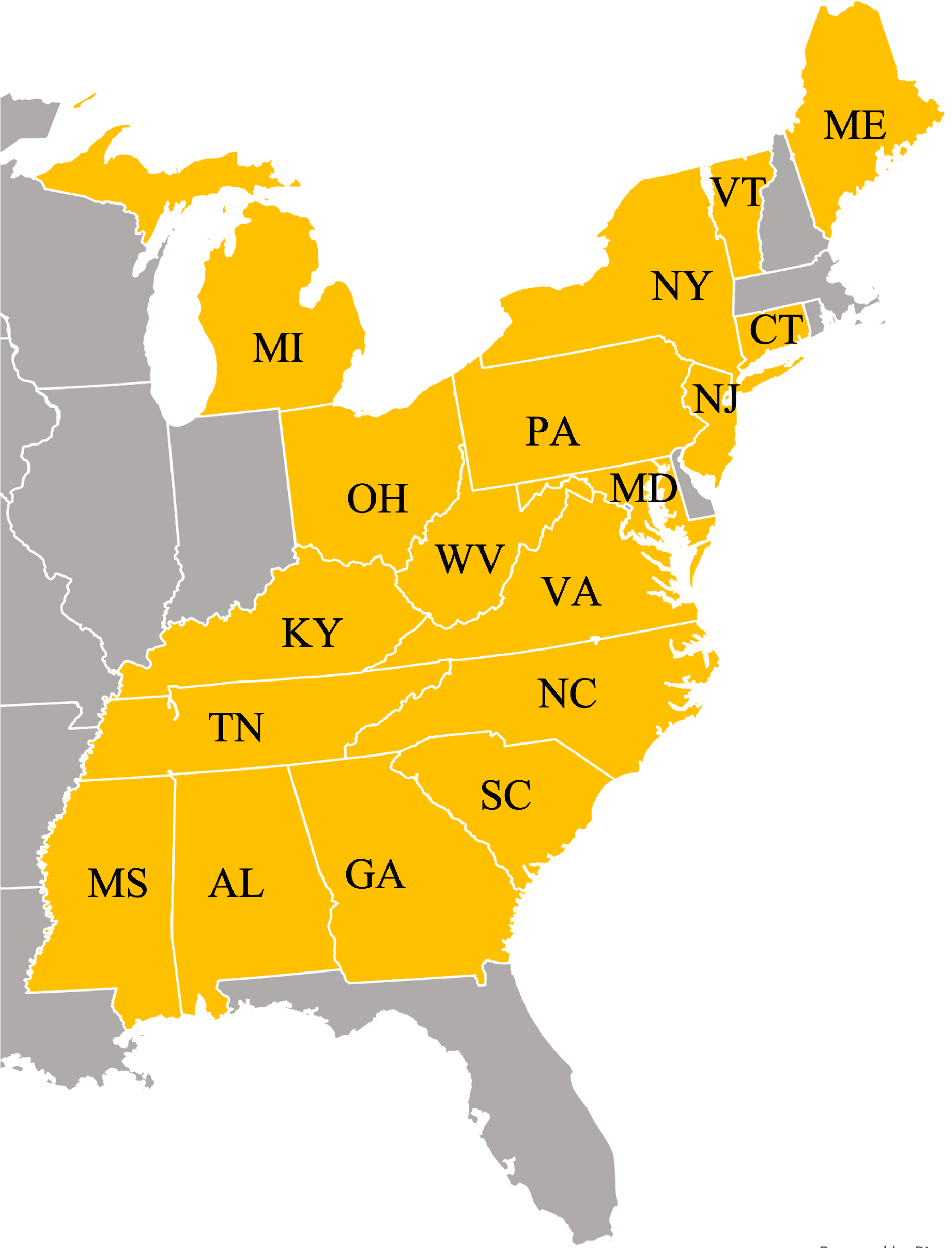}}
\subfigure[Workflow]{\label{fig2}\includegraphics[width=80mm]{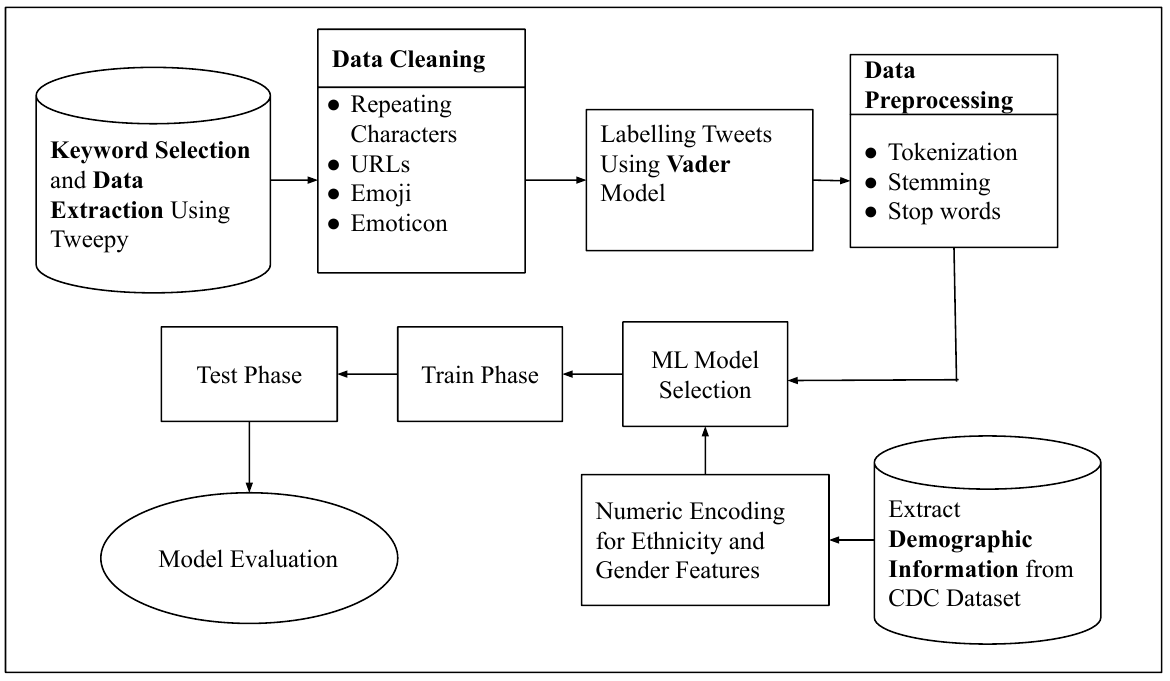}}
\caption{Selected States to Collect Tweets and Overall Workflow.}\label{fig1a}
\end{figure}

One of NLP’s biggest challenges is finding the appropriate dictionaries for the specific problem. Thus we developed a suitable NLP dictionary for CVD based on current literature. We choose some clinical and psychological risk factors as keywords causing CVD \cite{b17}, which people might use in their tweets. Then, we employed dictionary-based analysis to define the psychological language correlates of CVDs using hybrid CNN-LSTM architecture and other ML algorithms. We have chosen these keywords for causing CVD, which are considered risk factors to cause CVD or somehow create the possibility to relate to CVD. Based on those keywords, we have collected tweets from eighteen US states, including thirteen Appalachian states \cite{b18}, namely Alabama (AL), Georgia (GA), Kentucky (KY), Maryland (MD), Mississippi (MS), New York (NY), North Carolina (NC), Ohio (OH), Pennsylvania (PA), South Carolina (SC), Tennessee (TN), Virginia (VA), and West Virginia (WV). Apart from these Appalachian states, five other nearby states, namely Michigan (MI), New Jersey (NJ), Vermont (VT), Maine (ME), and Connecticut (CT), were chosen to extract tweets from those regions based on our selected keywords. Fig.~\ref{fig1} depicts the map of the selected states.

In this study, a state-wise CDC dataset from the US Department of Health \& Human Services was collected for 2019-2020\cite{b25}, containing 124 indicators related to chronic diseases. We applied the same ML models to a CDC dataset containing patients' demographic information in those eighteen states. The dataset contained demographic information like gender, race, ethnicity, location, diagnosis year, and disease type. Considering CVD as a class `1' and the other diseases as a class `0' throws a binary classification problem where we must classify individuals with CVD based on demographic information. The CDC dataset says the number of people who have already developed CVD in those states for that given timeframe. On the other hand, the Twitter-based dataset indicates the number of individuals who might develop CVD or already have CVD in those states for that timeframe. Overall, this paper has focused on Twitter being a great source in such predictions to assist public health practitioners.

\section{Related Works}
Previously, researchers employed NLP to track distinct diseases, i.e., flu, H1N1 influenza, depression, etc. \cite{b9,b10,b11,b12}. With the help of multi-source search engine data, Su et al. produced an accurate influenza forecast in regions with erratic seasonal influenza trends \cite{b13}. In pilot research, 135,000 tweets were collected during the Swine Flu pandemic in a week to extract communications and detect outbreaks from Twitter data \cite{b14}. Besides, Twitter data was utilized for mining public health topics in different studies \cite{b15,b16}.

In \cite{b4}, the researchers tried to find a connection between the language used on Twitter and heart disease mortality rates at the county level. By analyzing the psychological language patterns in tweets from over 1,300 U.S. counties, they discover that counties with more negative and less positive language on Twitter tend to have higher heart disease mortality rates. This research suggests that analyzing social media language can provide insights into public health outcomes, enabling the prediction of heart disease mortality and potentially informing targeted interventions (See counterclaim at \cite{b27}). 

In \cite{b5}, the researchers tried to investigate the use of Twitter as a data source for CVD research in response to the counterclaim. They analyzed a large sample of tweets to identify CVD-related keywords and hashtags. The study finds a significant correlation between the frequency of these CVD-related terms on Twitter and established risk factors and behaviors associated with cardiovascular health. This suggests that Twitter data has the potential to provide valuable insights into public discussions and behaviors related to CVD. The study highlights the utility of social media platforms like Twitter as a supplementary data source for monitoring and understanding cardiovascular health at a population level, which can inform public health interventions and targeted messaging for promoting cardiovascular well-being.

In \cite{b4} and \cite{b5}, the authors tried to find the correlation with CVD regarding demographic information and tweets. In our work, we tried to approach it differently. We developed our dictionary with keywords related to CVD to extract tweets. We used VADER for sentiment analysis to find the polarity. Based on the polarity of the sentiment, we assigned labels to the users with two classes, one with the risks of developing potential CVD and the other not developing any risk of CVD. In contrast to these two papers, we tried to assess this evaluation at the state level comprising eighteen states instead of the county level. Moreover, the timeframe of our collected tweets was also different. We are trying to denote that words expressed on social media can bear different meanings, and based on the risk factors of CVD, we can identify persons who have been developing the CVD.
\section{Methodological Framework}
Fig.~\ref{fig2} can be denoted as the methodological framework for this work as a work flow:

\subsection{Tweet-Based Approach}
\subsubsection{Keywords Selection and Dataset Collection}
Language analysis techniques for interpreting psychological states have a long history. Dictionary lists of terms connected to various structures are used in conventional methods (e.g., sad, glum, and crying are part of a negative-emotion dictionary). We have used the psychological and clinical keywords used in tweets to identify the persons with potential CVD risk. The keywords we have chosen are based on several categories. Anesthesia, angiogram, cardiologist, echocardiogram, heart attack, heart failure, hypertension, and chest pain are some of the most commonly used cardiac terminologies, according to Columbia Heart Surgery \cite{b17}. Smoking and cholesterol are two of the most known risk factors for heart-related diseases as per CDC \cite{b19}. Stress is considered to be a key psychological factor for causing CVD \cite{b20}, and alcohol use in excessive amounts can be a leading reason to cause CVD, according to John Hopkins Medicine \cite{b21}. More keywords or terminologies can be selected based on different criteria, but we decided to stick to twelve as a standard number. The keywords are then used for extracting the data from Twitter to analyze and feed into the model for identifying or predicting persons with potential CVD risks. Tweepy \cite{b22} is used to collect data from the Twitter API. By obtaining a developer account and the necessary access keys, tweets can be extracted based on specific keywords, locations, and periods. In this study, 269,969 tweets were collected over three years (2019 to 2021) from eighteen US states using twelve keywords. Collecting Twitter data using a specific set of keywords is indeed a common practice. However, the novelty of our study lies precisely in curating these CVD-related keywords. While traditional studies may rely on a broader range of CVD-related terms, our method stands out because it uses a carefully selected set of keywords that capture social media users' complex language patterns when discussing their cardiovascular disease experiences. Our keyword dictionary contains more than just formal clinical terms for CVD. We also included everyday language and lifestyle-related phrases often found in Twitter discussions about CVD. This allows our model to capture a broader and more realistic picture of how people talk about CVD on social media. 

\begin{figure}
\centering     
\subfigure[Class Distribution]{\label{fig:class}\includegraphics[width=50mm]{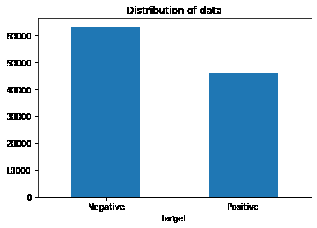}}
\subfigure[CNN-LSTM Architecture]{\label{fig:cnn}\includegraphics[width=70mm]{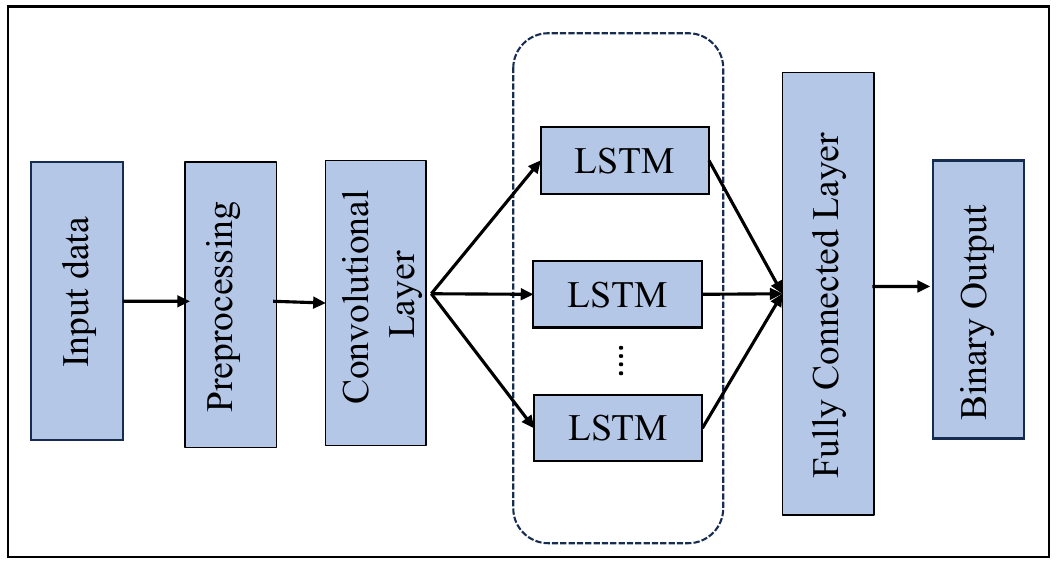}}
\caption{Twitter Data Class Distribution and CNN-LSTM Hybrid Architecture.}\label{fig3a}
\end{figure}

\subsubsection{Dataset Preprocessing and Polarity Identification}
Dataset preprocessing in NLP is a crucial step involving cleaning and formatting data before using it in ML or DL models. Preprocessing includes tokenizing, lowercasing, removing stop words and punctuation, and stemming. This study uses the VADER model to analyze sentiment in tweets and assess the potential risk of CVD. A threshold of -0.30 is set to identify positive sentiments related to CVD risks. Users above the threshold are labeled as `1' for potential risk, while those below are labeled as `0' for no risk \cite{b23, b24}. Model performance is evaluated based on sentiment analysis results, and the class distribution indicates sentiment labels (`1' for positive and `0' for negative). Fig.~\ref{fig:class} shows the class distribution after applying the VADER model. The dataset is split into a 70\%-30\% ratio for training and testing after labeling based on VADER analysis. In our study, we decided to use the VADER model with a threshold of -0.30 based on our preliminary work with a smaller dataset. We observed that this threshold closely aligned with our manual labeling process. It is important to note that this choice was made empirically, and we applied this threshold to label our larger training dataset. However, our main objective was not to assert that the VADER model with this specific threshold is universally the best option. Instead, we intended to use this labeled dataset as a reference for assessing and comparing the predictive performance of various ML and DL models in predicting CVD risks.
\subsubsection{Performance Evaluation with ML and DL}
This work utilizes a combination of ML and DL models. A hybrid CNN-LSTM model is constructed for the DL approach, as shown in Fig.~\ref{fig:cnn}. Conventional ML algorithms such as BNB, SVM, LR, and CatBoost are employed. Labels are assigned based on sentiment polarity identified by VADER, with `1' indicating positive sentiment and potential CVD risks and `0' indicating negative sentiment and no CVD risks. The dataset is split into training and test sets, with the ML and DL models trained on the training set and evaluated on the test set. The aim is to predict individuals with CVD risk based on sentiment expressed on Twitter. Performance evaluation metrics, including test accuracy, precision, recall, F1, MCC, and CK scores, are used to assess the performance of the classifiers.

\subsection{CDC Dataset-Based Approach}
Demographic information, including gender and ethnicity, was encoded numerically. One-hot encoding was used for the years 2019 and 2020. The focus was on predicting individuals with CVD. The dataset was balanced using the SMOTE technique \cite{b26}. Four ML models (BNB, SVM, LR, CatBoost) and a CNN-LSTM hybrid DL model were used. The dataset was split at 70\%-30\% ratio for training and testing. Limited demographic features were a challenge despite the large number of patients.

\section{Results Analysis}

\begin{figure}
\centering     
\subfigure[Test Accuracy]{\label{fig:s1}\includegraphics[width=60mm]{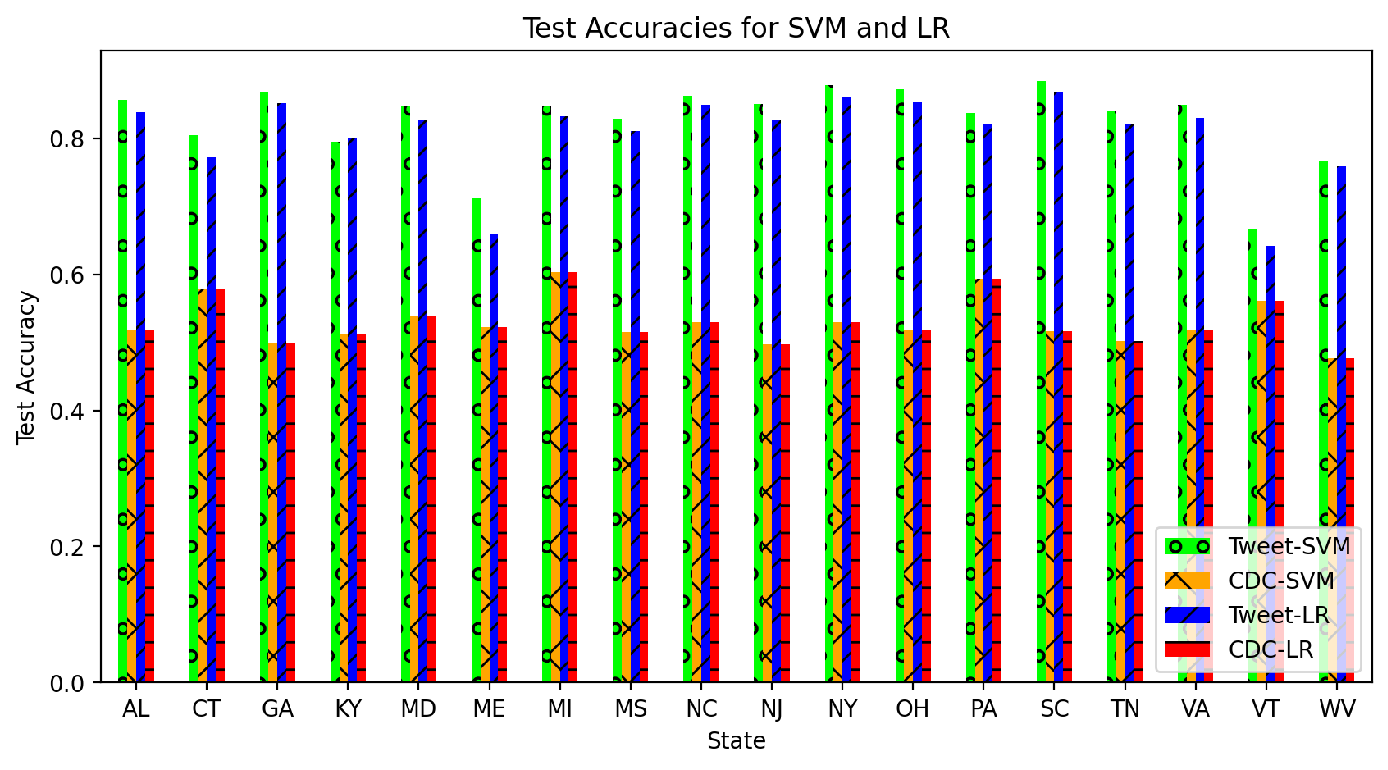}}
\subfigure[Ratio]{\label{fig:sratio}\includegraphics[width=60mm]{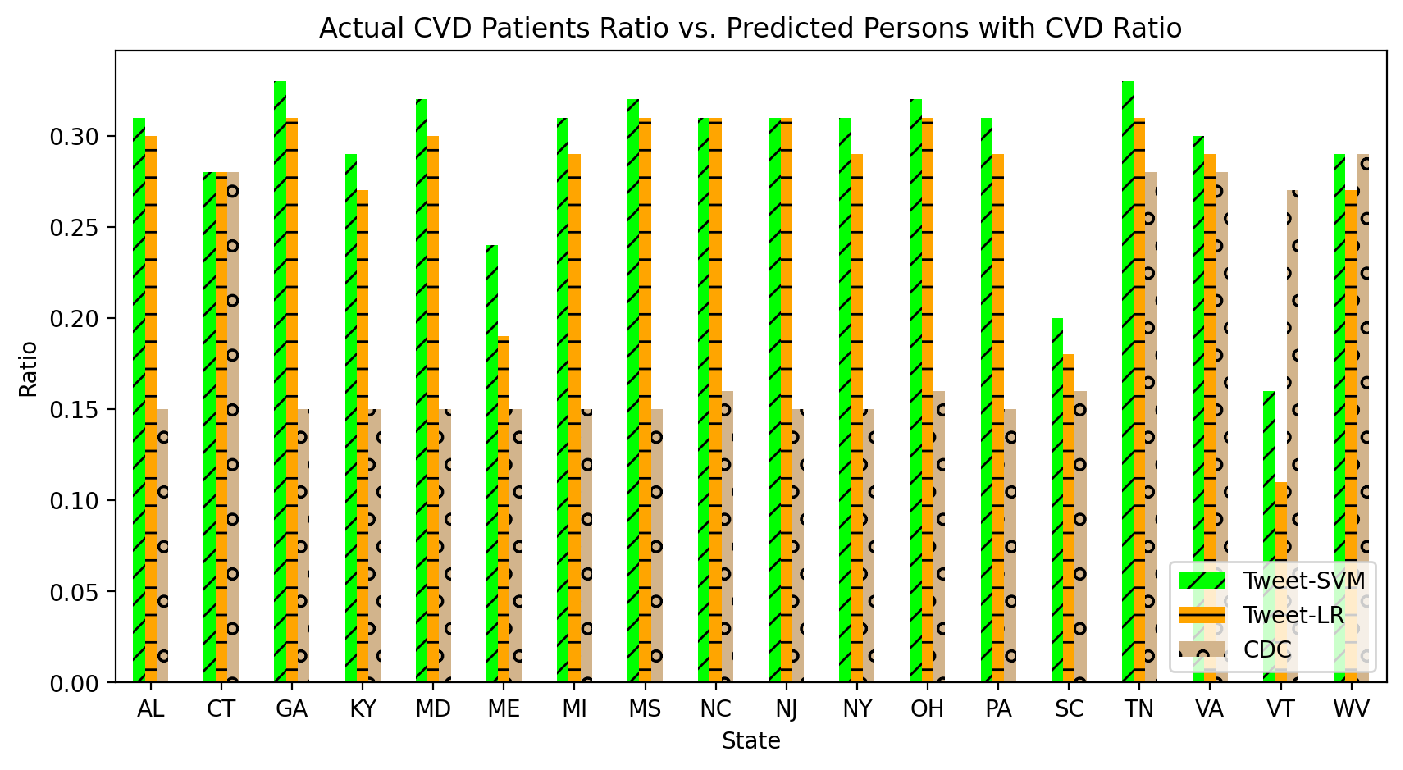}}
\caption{Results for SVM and LR.}\label{figsvm}
\end{figure}

Table~\ref{tab:1} shows the results for the overall Twitter dataset, consisting of tweets from eighteen states over three years. The CNN-LSTM hybrid model achieved 77.51\% test accuracy. SVM ranked highest with 88.75\% accuracy, followed by LR with 87.82\%. BNB and CatBoost achieved test accuracies of 74.55\% and 76.67\%, respectively. Table~\ref{tab:2} presents the results for the CDC dataset, where LR ranked highest with 58.03\% test accuracy, followed by BNB with 57.93\%. CNN-LSTM, SVM, and CatBoost achieved test accuracies of 57.64\%, 57.55\%, and 57.42\%, respectively. Comparative performance evaluations were conducted between the Twitter and CDC datasets, indicating that the Twitter-based dataset generally outperforms the CDC dataset. The study aims to establish the effectiveness of using NLP and ML with Twitter data to identify individuals at risk of CVD, which proves more effective than relying on demographic information alone. Fig.~\ref{fig:s1} illustrates the state-wise comparative performances of SVM and LR for Twitter and CDC datasets regarding test accuracy as the best two classifiers for Twitter dataset. Ratios comparing actual and predicted CVD prevalence based on CDC and Twitter data offer insights into the potential effectiveness of using social media data for predicting and estimating CVD prevalence which is outlined in Fig.~\ref{fig:sratio}. While these findings suggest the utility of social media data for public health research, further analysis and validation are required to ensure the reliability and generalizability of predictions derived from Twitter data.

\begin{table}[H]
\caption{Performance Evaluation for the Overall Twitter Based Dataset}
\label{tab:1}
\centering
\begin{tabular}{| p{2cm} | p{2.5cm} | p{1.6cm}| p{1.1cm}| p{0.8cm}| p{1.1cm}| p{1cm}|}
\hline
\textbf{Model} & \textbf{Test Accuracy} & \textbf{Precision} & \textbf{Recall} & \textbf{F1} & \textbf{MCC} &\textbf{CK}\\
\hline
CNN-LSTM & 77.51\% & 0.75 & 0.68 & 0.72 & 0.53 & 0.53\\
\hline
BNB & 74.55\% & 0.84 & 0.48 & 0.61 & 0.48 & 0.44\\
\hline
SVM & 88.75\% & 0.87 & 0.86 & 0.86 & 0.77 & 0.77\\
\hline
LR & 87.82\% & 0.85 & 0.86 & 0.85 & 0.75 & 0.75\\
\hline
CatBoost & 76.67\% & 0.73 & 0.71 & 0.72 & 0.52 & 0.53\\
\hline
\end{tabular}
\end{table}

\vspace{-3.2\baselineskip}
\begin{table}[H]
\caption{Performance Evaluation for the CDC Dataset}
\label{tab:2}
\centering
\begin{tabular}{| p{2cm} | p{2.5cm} | p{1.6cm}| p{1.1cm}| p{0.8cm}| p{1.1cm}| p{1cm}|}
\hline
\textbf{Model} & \textbf{Test Accuracy} & \textbf{Precision} & \textbf{Recall} & \textbf{F1} & \textbf{MCC} &\textbf{CK}\\
\hline
CNN-LSTM & 57.64\% & 0.63 & 0.36 & 0.45 & 0.17 & 0.15\\
\hline
BNB & 57.93\% & 0.67 & 0.31 & 0.42 & 0.19 & 0.16\\
\hline
SVM & 57.55\% & 0.61 & 0.41 & 0.49 & 0.16 & 0.15\\
\hline
LR & 58.03\% & 0.62 & 0.39 & 0.48 & 0.17 & 0.16\\
\hline
CatBoost & 57.42\% & 0.61 & 0.42 & 0.50 & 0.16 & 0.15\\
\hline
\end{tabular}
\end{table}
\vspace{-2.5\baselineskip}
\section{Acknowledgment}
This work was supported in part by grants from the US National Science Foundation (Award \#1920920, \#2125872).
\section{Conclusions}
This study collected tweets from eighteen US states using a new dictionary. Before conducting sentiment analysis, preprocessing techniques like stemming and lemmatization were applied to the tweets. The VADER model was then used to determine the polarity of the tweets (negative or positive) and assign labels for ML and DL models. ML models such as BNB, SVM, LR, and CatBoost were utilized, along with the CNN-LSTM hybrid model for DL. The SVM achieved the highest test accuracy of 88.75\%, followed by LR with 87.82\% accuracy. CNN-LSTM achieved 77.51\% test accuracy. The study demonstrates that NLP and ML can effectively use Twitter data to predict individuals with potential CVD risk, with sentiment analysis aiding in label creation. Results were compared with the CDC dataset with demographic information to validate the claim. Additionally, performance evaluation parameters can be incorporated to assess the models further. State-wise performance was comparatively better for tweets than demographic information for the best two classifiers. This indicates that tweets are a valuable resource for predicting or classifying individuals at risk of CVD compared to demographic data. 

%
%
%

\bibliographystyle{splncs04}
\bibliography{References}

\end{document}